\documentclass[nohyperref]{article}

\usepackage{microtype}
\usepackage{graphicx}
\usepackage{subfigure}
\usepackage{booktabs} 
\usepackage{sidecap}

\usepackage{hyperref}

\usepackage[accepted]{icml2022}

\usepackage{amsmath}
\usepackage{amssymb}
\usepackage{mathtools}
\usepackage{amsthm}

\usepackage[capitalize,noabbrev]{cleveref}

\theoremstyle{plain}

\theoremstyle{definition}

\theoremstyle{remark}

\usepackage[textsize=tiny]{todonotes}

\newcommand{\encftrs}{\Phi_{\textrm{enc}}}

\newcommand{\heatmapftrs}{\Phi_{\textrm{heatmap}}}
\newcommand{\dbypass}{\Phi_{\textrm{bypass}}}
\newcommand{\dgraph}{\Phi_{\textrm{graph}}}
\newcommand{\dglimpse}{D_{\textrm{glimpse}}}
\newcommand{\dupsample}{D_{\textrm{upsample}}}

\newenvironment{packed_enumerate}{
\begin{enumerate}
  \setlength{\itemsep}{1pt}
  \setlength{\parskip}{0pt}
  \setlength{\parsep}{0pt}
}{\end{enumerate}}

\icmltitlerunning{Unsupervised Image Representation Learning with Deep Latent Particles}

\begin{document}

\twocolumn[
\icmltitle{Unsupervised Image Representation Learning with Deep Latent Particles}




\begin{icmlauthorlist}
\icmlauthor{Tal Daniel}{yyy}
\icmlauthor{Aviv Tamar}{yyy}
\end{icmlauthorlist}

\icmlaffiliation{yyy}{Department of Electrical and Computer Engineering, Technion - Israel Institute of Technology, Haifa, Israel}

\icmlcorrespondingauthor{Tal Daniel}{taldanielm@campus.technion.ac.il}

\icmlkeywords{Machine Learning, ICML, Deep Learning, Keypoint, VAE, variational, autoencoder, unsupervised}

\vskip 0.3in
]



\printAffiliationsAndNotice{}  

\begin{abstract}
We propose a new representation of visual data that disentangles object position from appearance. Our method, termed Deep Latent Particles (DLP), decomposes the visual input into low-dimensional latent ``particles'', where each particle is described by its spatial location and features of its surrounding region. To drive learning of such representations, we follow a VAE-based approach and introduce a prior for particle positions based on a spatial-softmax architecture, and a modification of the evidence lower bound loss inspired by the Chamfer distance between particles. 
We demonstrate that our DLP representations are useful for downstream tasks such as unsupervised keypoint (KP) detection, image manipulation, and video prediction for scenes composed of multiple dynamic objects. In addition, we show that our probabilistic interpretation of the problem naturally provides uncertainty estimates for particle locations, which can be used for model selection, among other tasks. Videos and code are available: \url{https://taldatech.github.io/deep-latent-particles-web/}.
\end{abstract}

\section{Introduction}
\label{sec:intro}

The spatial positions of various parts of an image contain useful information for decision making. Examples include the positions of objects in video games~\cite{smirnov2021marionette}, rigid object poses in robotic manipulation~\cite{byravan2017se3}, and image landmarks for human pose estimation~\cite{jakab2018unsupervised}. This observation motivates us to seek image representations where position is disentangled from other visual properties of objects in the scene.

In different image types, however, the definition of objects and their positions may vary (e.g., contrast the position of facial parts such as  eyes and nose with the position of computer game sprites), leading us to pursue an unsupervised approach, where representations are data-dependent. In particular, we follow the generative approach -- learn to reconstruct an image from its disentangled latent representation~\cite{burgess2019monet}.

This problem has recently gained attention in the variational autoencoding (VAE) literature, 
motivated by the observation that the typical single-vector representation of the entire scene has issues in scenes with multiple, varying number of objects. A common remedy is decomposing the scene into a pre-determined number of objects~\citep{burgess2019monet, watters2019cobra, kipf2019cswm}, and then learning a separate representation for each object. The main caveats with these methods are their strong assumptions on the number of objects, and their complexity, as the object discovery process is usually iterative and initialization-dependent~\citep{burgess2019monet, kipf2019cswm}.

An alternative representation is based on object landmarks, or \textit{keypoints}, an idea that dates back to classical computer vision~\citep{sift99}. A keypoint (KP) representation is an unordered set of geometrical points (i.e., $(x, y)$ locations in 2D scenes and $(x,y,z)$ in 3D scenes). Several recent works investigated learning keypoints with deep learning methods~\citep{jakab2018unsupervised, thewlis2017unsupervised1, zhang2018kp, lorenz2019unsupervised, dundar2021unsupervised}, exploiting the translational-invariance and spatial-locality of convolutional-based architectures. Predominantly, \citet{jakab2018unsupervised} apply the  spatial-softmax over feature maps extracted from a convolutional neural network (CNN) to determine the location of the keypoints. This idea became a building block in other methods that use KP for downstream tasks~\citep{kulkarni2019transporter, gopalakrishnan2020permakey, boney2021keyq, he2021latentkeypointgan}, and has proven to be a promising alternative to the single-vector representation.

In this work, we propose a new representation method that draws inspiration from both KP and VAEs. Our idea is to \textit{view the keypoints themselves as the latent space of the VAE}. To sufficiently capture image information, we accompany each KP with a set of features to describe the content in its vicinity, and refer to it as a ``particle''. Our method, termed Deep Latent Particles (DLP), inherits favorable traits from the VAE approach, such as a natural decoding scheme for reconstructing the image, and an uncertainty estimate for the particles that is based on the probabilistic interpretation of the latent variables. Meanwhile, 
as particles are jointly encoded and decoded, DLP does not require iterative inference, and can work with a larger number of objects ($>30$) than prior VAE-based methods. Our method is also more flexible than recent patch-based approaches~\citep{smirnov2021marionette, lin2020space}, which anchor objects to the center of the patch -- our encoder allows particles to be freely located on the canvas, and for multiple particles to jointly model a single object (e.g., a large object).

Key for making our method work are two novel ideas. The first is to add a prior on particle positions that is based on the spatial-softmax architecture. The second is to learn the posterior of particle positions based on a modification of the evidence lower bound (ELBO) that is inspired by the Chamfer distance between particles, which we term the \textit{Chamfer KL}. Building on these two ideas, we propose a VAE-inspired model where particles act as latent variables.

We demonstrate DLP on datasets that contain scenes with and without explicit objects. Our results show that the learned latent space effectively disentangles position from appearance. For example, when trained on CelebA~\citep{liu2015faceattributes} data, moving the particle located on the nose only controls the nose region in the output image.
Importantly, we show that incorporating the particle uncertainty can benefit downstream tasks such as supervised landmark regression, where we demonstrate state-of-the-art results. Finally, we demonstrate that our method can be used to manipulate scenes with multiple objects by changing the location of the particles, and how this idea can be used for video prediction, by training a graph neural network (GNN) to predict the change in the particles.

Our contributions are summed as follows: (1) we propose a new unsupervised particle-based latent representation, trained with a novel modification of the VAE loss function based on the Chamfer distance for a set of points; (2) we show that our method is capable of extracting objects and their masks from multiple-object scenes without any supervision; (3) we experiment with various image datasets, showing the method's applications in keypoint discovery, image manipulation and video prediction; and (4) we demonstrate the benefits of incorporating the learned uncertainty information for model selection in the task of KP regression.

\section{Related Work}
\label{sec:rel_work}
Our work is inspired by ideas from unsupervised keypoint detection and unsupervised scene decomposition to objects.

\textbf{Unsupervised keypoint detection:} \citet{thewlis2017unsupervised1} represent the structure of an object as a set of keypoints learned via image deformations under equivariance constraints. 
\citet{zhang2018kp} proposed an autoencoding-based landmark discovery approach with a constrained bottleneck, to improve upon \citet{thewlis2017unsupervised1}. The method does not require pairs of images, but introduces several equivariance and separation constraints, making it more complex. \citet{jakab2018unsupervised} proposed KeyNet\footnote{While \citet{jakab2018unsupervised} did not officialy name their method, it is often referred to as KeyNet \citep[e.g.,][]{gopalakrishnan2020permakey}.} to learn KP using a tight bottleneck. KeyNet is simpler -- the constraints of \citet{zhang2018kp} are removed, but it requires pairs of images (video frames or augmented images). KeyNet outperformed \citet{zhang2018kp} by a large margin.
Transporter \citep{kulkarni2019transporter} extends KeyNet by learning to transport image features between two frames for tracking objects and object parts across long time-horizons.
The aforementioned methods learn deterministic keypoints, without uncertainty information as our DLP. Moreover, we show that in the single-image setting we outperform \citet{zhang2018kp} without requiring their constraints, and outperform \citet{jakab2018unsupervised} in the image-pairs setting as well.

Recently, several methods  complimented KP discovery with modules that model object parts \citep{lorenz2019unsupervised, dundar2021unsupervised}. These methods disentangle shape and appearance by using various types of augmentations, and introduce components that mask foreground and background. 
This additional complexity proved useful, demonstrating outstanding results in various computer applications. Our method is simpler, as it uses standard reconstruction loss terms, and provides a natural uncertainty measure for the keypoints. Additionally, as DLP represents the whole image using the latent particles, it allows to explicitly control the generated scene, unlike previous methods that rely on feature maps from the original image, and can therefore only modify small local regions around the KP.

\textbf{Unsupervised object-centric representations:} unsupervised discovery of objects in scenes has mainly relied on sequential inference of objects, where in each iteration a new part of the input is attended to, or patch-based inference, where each patch can contain an object that needs to be represented. 
AIR~\citep{eslami2016attend}, SQAIR~\citep{kosiorek2018sequential}, R-SQAIR~\citep{stanic2019rsqair} and SPAIR~\citep{crawford2019spair} are based on sequential inference of objects and explicitly representing objects as ‘what’, ‘where’, and ‘presence’ latent variables, where the latter also adds a `depth' variable. APEX~\citep{wu2021apex} leverages a similar approach but exploits temporal information in videos for better performance.
These models are limited to a moderate number of objects, and typically struggle with modelling the scene's background. In contrast, our inference happens all at once, and the `where' representation is replaced with a spatial prior over keypoint locations.  
MONet~\citep{burgess2019monet} uses a sequential attention mechanism to allocate objects to slots, while Slot Attention~\citep{locatello2020object} replaces MONet's multi-step procedure with a single step using iterated attention between slots. IODINE~\citep{greff2019iodine} adds iterative refinement to objects, exhibiting similar performance to MONet, but requires more memory and is arguably more complex. GENESIS~\citep{engelcke2019genesis} and GENESISv2~\citep{engelcke2021genesis2}  build on MONet and IODINE and introduce a generative model that captures relations between scene components with an autoregressive prior. Our method does not contain autoregressive components or iterative inference.

Finally, MarioNette~\citep{smirnov2021marionette} and SPACE~\citep{lin2020space} are non-sequential patch-based approaches. SPACE factors patches into ‘what’, ‘where’, `depth', and ‘presence’ in parallel and is thus more scalable than the aforementioned methods, but has a tendency to embed objects in the background. MarioNette takes a different approach and learns a deterministic dictionary of objects or sprites that can appear in a scene. However, the dictionary approach is discrete in nature and is limited to objects seen during training. Our method is also non-sequential, but is not limited to patches or a static dictionary of learned objects.

\textbf{Latent video prediction:} recent advances in generative modelling~\citep[e.g.,][]{razavi2019vqvqae, karras2020analyzing, daniel2020soft} have inspired a large body of video prediction methods that employ prediction in a learned latent space~\citep{minderer2019unsupervised, wu2021greedy, walker2021predicting, villegas2019high, yan2021videogpt}. \citet{walker2021predicting, yan2021videogpt} model a sequence of discrete latent variables in the latent space of a vector-quantized VAE, while \citet{wu2021greedy} and \citet{villegas2019high} focus on scaling-up latent autoregressive video prediction, the first via hierarchical VAEs and the latter via large stochastic recurrent neural networks~(RNNs). V-CDN~\citep{li2020causal} performs video prediction of physical interaction by building a causal graph from keypoints learned with Transporter~\citep{kulkarni2019transporter}.
Closely related to our work, \citet{minderer2019unsupervised} uses KeyNet~\citep{jakab2018unsupervised} to learn keypoints, and propose a variational RNN to model stochastic dynamics. We empirically compare with this approach, and report improved performance on datasets with varying number of objects, which we attribute to the GNN in our method. However, the 
dynamics model in \citet{minderer2019unsupervised} is orthogonal to our work, and can potentially be used with our DLP representation as well.

\section{Background}
\label{sec:bg}
\textbf{Variational Autoencoders (VAEs):} VAEs~\citep{kingma2014autoencoding} learn an approximate model of data $p_{\theta}(x)$ using variational inference by maximizing the evidence lower bound (ELBO), which states that for any approximate posterior distribution $q(z|x)$: 
\begin{equation}\label{eq:ELBO_def}
\begin{split}
    \log p_\theta(x) &\geq \mathbb{E}_{q(z|x)}\ \left[\log p_\theta(x|z)\right] - KL(q(z|x) \Vert p(z)) \\
    & \doteq ELBO(x), \raisetag{1.5em}
\end{split}
\end{equation}
where the Kullback-Leibler (KL) divergence is
$KL(q(z|x) \Vert p(z)) = \mathbb{E}_{q(z|x)} \left[ \log \frac{q(z|x)}{p(z)} \right]$. Typically, Gaussian distributions are used to model the approximate posterior $q_{\phi}(z|x)$, likelihood $p_{\theta}(z|x)$, and prior $p(z)$. The approximate posterior $q_\phi(z|x)$ is also known as the \emph{encoder}, while $p_\theta(x|z)$ is termed the \emph{decoder}. The ELBO can be maximized using the \textit{reparameterization trick}, and in what follows, the term \textit{reconstruction error} refers to $\log p_\theta(x|z)$.

\textbf{KeyNet:} The objective in~\citet{jakab2018unsupervised} is to produce a set of $K$ 2D coordinates (a.k.a.~keypoints/landmarks) $y=(u_1,...,u_K), u_k\in \Omega$ for a given image, where $\Omega = [-1,1]^2$ denotes the normalized space of 2D positions in the image.
Consider an image $x \in \mathcal{R}^{H \times W \times 3}$. To extract keypoints, an encoder network (CNN) outputs $K$ feature maps $S_u(x;k) \in \mathcal{R}^{H' \times W'}$, $k=1,...,K$.
The keypoint $u_k$ is finally generated from $S_u(x;k)$ using a spatial softmax (SSM) layer~\cite{finn2016deep}.

\textbf{Gaussian Heatmaps:} To backpropagate through the keypoint positions, \citet{jakab2018unsupervised} broadcast each keypoint $u_k$ into a Gaussian-like 2D heatmap centered around $u_k$ with a small and constant standard deviation $\sigma$: $\Phi_u(x;k) = \exp{\left(-\frac{1}{2\sigma^2}|| u - u_k(x)||^2\right)} .$ These maps are then used as part of the image reconstruction process.

\textbf{Chamfer distance:} the distance between point clouds $S_1$ and $S_2$ of arbitrary size can be calculated with the Chamfer distance: $ d_{CH}(S_1, S_2) = \sum_{x \in S_1}\min_{y \in S_2}||x-y||_2^2 + \sum_{y \in S_2}\min_{x \in S_1}||x-y||_2^2.$
\section{Method}
\label{sec:method}

Our objective is to design a VAE for images $x \in \mathcal{R}^{H \times W \times C}$ where the latent representation is structured as a set of particles $z  \in \mathcal{R}^{K \times (2+d)}$, where $K$ is the number of particles, the first two components of each particle contain information about the positions of objects, and $d$ additional features encode information about object appearance. Explicitly, we denote $z =[z_p, z_a]$, where $z_p \in \mathcal{R}^{K \times 2}$ and $z_a  \in \mathcal{R}^{K \times d}$ denote the position and appearance components, respectively.

\begin{center}
\begin{figure*}
    \centering
    \includegraphics[width=0.9\textwidth]{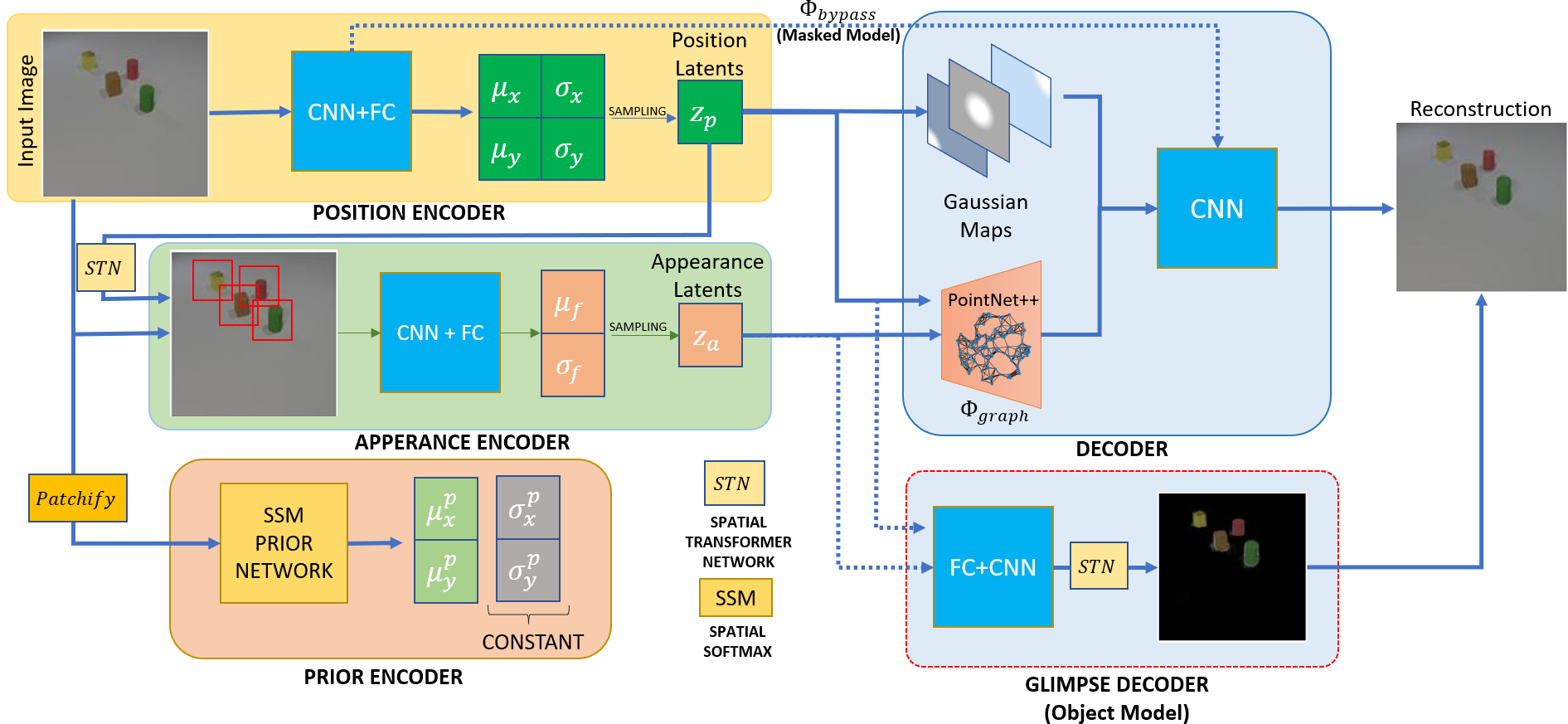}
    \vspace{-1.0em}
    \caption{DLP architecture. Image is processed by the position encoder to produce the posterior probability of latent particle positions. These positions are used to extract glimpses from the original images using a STN, which are then processed by an appearance encoder to produce the appearance features for each particle. The input image (or an augmented view of it) is also processed by the prior network, producing keypoint proposals via SSM. To reconstruct the image, the particles are (1) transformed to differentiable Gaussian heatmaps and (2) go through a PointNet++ to produce feature maps $\Phi_{graph}$. For the \texttt{Masked} model, the heatmaps are used as binary masks to combine local regions from $\Phi_{graph}$ with bypass features from the encoder. 
    For the \texttt{Object} model, a separate glimpse decoder is used to decode RGBA patches, which are then combined with feature maps $\Phi_{graph}$ to produce the output image. See text for full details.}
    \label{fig:vae_arch}
    \vspace{-1em}
\end{figure*}
\end{center}

There are several challenges in doing so. 
The first is how to disentangle the positional information from other content in the particle features, without sacrificing reconstruction quality. Our idea is to exploit the probabilistic interpretation of the VAE, and view each particle position as coming from a distribution, where its prior is given by the standard method of \citet{jakab2018unsupervised}, which we already know produces reasonable keypoints. The posterior neural network, in contrast, is not restricted to any particular structure, to allow maximal expressiveness for accurately reconstructing the scene. The KL term in the VAE loss, which forces the posterior to be close to the prior, will drive the posterior to have positional meaning. This idea brings about another challenge -- the prior can generate more keypoints than required by the posterior, so we need a method to enforce similarity between two sets of keypoint of different sizes. Finally, we need to also encode (and decode) the appearance information \textit{around} the predicted keypoints. For this we propose the appearance encoder, which encodes `glimpses' of the image around the keypoint in a differentiable way. In the following, we explain each of these components in detail. The model is illustrated in Figure \ref{fig:vae_arch}.

\vspace{-3em}
\subsection{Patch-wise Conditional Prior}
In the standard VAE setting, the prior distribution is fixed and usually set to $p(z) = \mathcal{N}(0, I)$. Here we consider a conditional setting~\citep{Sohn2015cvae}, where we explicitly learn a prior for the keypoints given an image $x$, $p_{\psi}(z|x) = p_{\psi}(z_p|x)\times p_{\psi}(z_a|x)$.\footnote{ \citet{Sohn2015cvae} show that such a conditional prior complies with the ELBO formulation.} We set $p_{\psi}(z_a|x)$ to the unit Gaussian prior $\mathcal{N}(0, I_d)$. The prior for the positions, however, requires more sophistication, as we describe next.

First, note that for computing the ELBO in Eq.~\eqref{eq:ELBO_def}, we do not need the explicit form of the prior, but only the KL divergence between the prior and approximate posterior. In the following, the prior distribution will be defined implicitly, through a set of prior keypoint proposals, and a particular KL divergence term to be described in Section \ref{subsec:chamfer_kl}.

To generate keypoint proposals, 
the input $x \in \mathcal{R}^{H \times W \times 3}$ is split into $K_p$ (in general $K_p\neq K$) patches of size $D \times D$ (in our experiments $D\in\{8, 16\}$), and for each patch, the prior network outputs a distribution for the coordinates of a single keypoint proposal. This distribution is Gaussian $\mathcal{N}(\mu_p, \sigma_p)$, where the standard deviation is chosen to be a fixed small constant $\sigma_p=0.1$, and the mean is the output of SSM, as described in Section~\ref{sec:bg}.
Therefore, for each image we obtain a set of $K_p$ unordered keypoint proposals.
In practice, as the set of proposals can grow large with the number of patches, we found it useful to consider only a subset of $L$ prior keypoints, where $L$ is a hyper-parameter and we term the set of prior keypoints \textit{keypoint proposals}. To filter out $L$ keypoints, we explored uniform sampling of $L$ from the set of $K_p$ proposals, and a heuristic where we keep the top-$L$ distant keypoints from the center of their respective center. The logic behind this heuristic is that applying SSM in smooth patches (e.g., a solid color background) will result in a keypoint in the center of the patch, which might be uninformative. Both filtering methods resulted in similar performance, with a slight advantage to the heuristic.

\subsection{Position Encoder and Appearance Encoder}
The encoder models the approximate posterior, $q_{\phi}(z|x)=q_{\phi}(z_p|x)\times q_{\phi}(z_a|x, z_p)$. The \textit{position encoder} $q_{\phi}(z_p|x)$ has a similar architecture to KeyNet~\citep{jakab2018unsupervised}: the input image is downsampled with convolutional layers ending with $K$ feature maps $\encftrs(x) \in \mathcal{R}^{H' \times W' \times K}$. 
Unlike KeyNet, however, we do not apply SSM on these feature maps, but flatten them and use a fully-connected (FC) layer to map to $K$ keypoints. The output of the FC layer is $\mu, \log(\sigma^2) \in \mathcal{R}^{K \times 2}$, the means and log variances of $K$ independent Gaussians that make up $q_{\phi}(z_p|x)$.

For the \textit{appearance encoder}, $q_{\phi}(z_a|x, z_p)$, we desire the features to encode the visual properties of the vicinity of each keypoint, in a differentiable manner.
To that end, we use a Spatial Transformer Network (STN~\citealt{jaderberg2015stn}), similarly to SPAIR~\citep{crawford2019spair}, to extract regions of size $S\times S$ from the original input at the locations specified by $z_p$, where the region size $S$ around the keypoint is a hyperparameter (in our work, $S \in \{16, 32\}$). These \textit{glimpses} go through 
a small CNN ending with a FC layer mapping to the parameters $\mu_f, \log(\sigma_f^2)\in \mathcal{R}^{d}$ of a Gaussian distribution for the features of each particle.

\subsection{Chamfer-KL Distance}\label{subsec:chamfer_kl} 
The use of a FC layer instead of SSM in the encoder gives the model additional freedom when choosing the ideal keypoint locations for reconstruction. 
Note, however, that the number of posterior keypoints $K$ does not necessarily equal $L$, the number of prior keypoint proposals from $p_{\psi}(z_p|x)$. 

We constrain the posterior to be close to the prior despite differences in the number and ordering of elements in each set, using a novel loss that we term the Chamfer-KL distance. 
Chamfer-KL views each point cloud as a set of Gaussian distributions and calculates the KL divergence between a keypoint in set $S_1$ and every keypoint in set $S_2$: $$ d_{CH-KL}(S_1, \!S_2)\! = \!\!\!\sum_{x \in S_1}\!\min_{y \in S_2}KL(x \Vert y) +\!\! \sum_{y \in S_2}\!\min_{x \in S_1}KL(x \Vert y). $$ Note that unlike the $L_2$ distance, the KL is asymmetrical and thus not a metric, a property which is maintained in the Chamfer-KL as $d_{CH-KL}(S_1, S_2) \neq d_{CH-KL}(S_2, S_1)$. 

\subsection{Decoder}
\label{subsec:kp_dec}
The decoder in a VAE maps the latent variable $z$ into an image. For our  particle-based latents, we found that different decoder architectures work better for different types of scenes, depending on the variation of the background in the data, and whether the scene is composed of many separated objects or not. We next describe three basic decoder components, and how to combine them for different scenes.

The basic decoder component is an upsampling CNN $\dupsample$ 
taking in input feature maps 
and outputting the reconstructed image $\Tilde{x} \in \mathcal{R}^{H \times W \times 3}$. The input feature maps of $\dupsample$ are comprised of a concatenation of Gaussian heatmaps $\heatmapftrs$ constructed from the coordinates of each particle as described in Section~\ref{sec:bg}, and feature maps. We explored two methods for  generating the feature maps, the \textit{graph component} $\dgraph$ and the  \textit{bypass component} $\dbypass$. 

Additionally, for multi-object scenes, we introduce a separate \textit{glimpse decoder} $\dglimpse$, taking in a single particle and outputting an RGBA patch of the surrounding region around the particle. The output of $\dupsample$ and the output of $\dglimpse$ can be stitched together to create the final reconstructed image. We next overview each component; a full technical description is in the Appendix.

\textbf{Graph component $\dgraph$:} The goal of this component is to use the particles for modelling the global structure of the scene. We create a KNN graph, where nodes are the particles, and edges are connected based on the Euclidean distance between particle positions. This graph is processed by a PointNet++~\citep{qi2017pointnet++}, which, using global max pooling, outputs feature maps $\Phi_p(x) \in \mathcal{R}^{H' \times W' \times M}$.\footnote{In all our experiments we arbitrarily chose $M=K$.}

\textbf{Bypass component $\dbypass$:} The goal of this component is to supply the background features that are not modelled by the particles.
We simply take the feature maps from the encoder $\encftrs(x)$.
A similar component was employed in \citet{jakab2018unsupervised}, but was used to model the whole scene without considering the features around the keypoints.

\textbf{Glimpse decoder $\dglimpse$:} This component reconstructs each object's appearance independently, as a combination of RGB values and a mask (alpha channel).  We use a fully-connected layer followed by a small upsampling CNN that takes in a single latent particle $z_i$ and decodes an RGBA patch of the surrounding region around the particle $\Tilde{x^p_i} \in \mathcal{R}^{S \times S \times 4}$. The decoded patches are positioned in the full $H \times W$ canvas according to their respective particle's position, using a STN.

Equipped with the decoder components defined above, we now describe two  decoder combinations that we explored:
\begin{packed_enumerate}
\vspace{-1em}
    \item \texttt{Masked} model: $\texttt{concat}[\heatmapftrs, \alpha \times \dgraph, (1-\alpha) \times \dbypass] \to \dupsample$. 
    The masks $\alpha$ are generated from the Gaussian heatmaps $\heatmapftrs$, and represent foreground. 
    \item \texttt{Object-based} model: $\alpha \times \dglimpse + (1-\alpha)\times (\texttt{concat}[\heatmapftrs, \dgraph] \to \dupsample)$. The masks $\alpha$ are given by the alpha channel of $\dglimpse$.\footnote{The careful reader may wonder why we used $\dgraph$ instead of $\dbypass$ for the background. Empirically, we found that $\dbypass$ can be too expressive, causing some of the objects to be represented by it instead of by $\dglimpse$.}
\end{packed_enumerate}
\vspace{-1em}
The \texttt{Object-based} model has a \textit{pure} bottleneck -- all information from the encoder to the decoder flows through the latent representation. In the \texttt{Masked} model, however, the bypass component  skips the bottleneck. As we show in our experiments, the pure bottleneck model allows the particles more control over the output image, while the \texttt{Masked} model excels in reconstruction quality.

\subsection{Training the Model}
\label{subsec:kp_vae_train}
The training procedure is based on maximizing the ELBO \eqref{eq:ELBO_def}.
Since all of our distributions are Gaussians, the KL divergence has a closed form solution. Similar to the $\beta$-VAE~\citep{higgins2017beta}, we multiply the KL and Chamfer-KL terms in the loss by hyperparameters $\beta_{\textrm{KL}}$ and $\beta_{\textrm{CKL}}$, respectively.
To obtain better image quality, we replace the typical MSE reconstruction error with a variant of the VGG perceptual loss~\citep{hoshen2019non} that calculates the $L_2$ distance between the extracted features instead of the $L_1$, similarly to \citet{jakab2018unsupervised}.

The model is optimized end-to-end by Adam~\cite{adam_14}, using the reparametrization trick, and is implemented in PyTorch~\citep{paszke2017automatic}. Extended implementation details can be found in the Appendix. Our code is available publicly\footnote{\url{https://github.com/taldatech/deep-latent-particles-pytorch}}.

\section{Experiments}
\label{sec:exp}
Our method produces latent particles to represent an input image. We design our set of experiments to answer the following questions: (1) does our method effectively disentangle position from appearance in various scene types; (2) how important are our novel Chamfer-KL and the conditional SSM prior components;
(3) what is the quality of our particles compared to other unsupervised keypoint discovery methods; 
(4) can our approach be used for downstream tasks such as image manipulation; and (5) can we exploit our probabilistic formulation to infer uncertainty estimates for the particles.

\subsection{Linear Regression on Face Landmarks}
\label{subsec:supervised_kp}
A standard quantitative evaluation of unsupervised keypoint discovery is based on the error in predicting annotated keypoints from the discovered keypoints. 

The common benchmark for this task uses the CelebA train set while excluding the MAFL~\citep{zhang2014facial} subset which includes annotations for 5 facial landmarks-- eyes, nose and mouth corners.
We train our \texttt{Masked}-model with a similar architecture and pre-processing as \citet{jakab2018unsupervised} on $128\times 128$ face images from CelebA~\citep{liu2015faceattributes}. For the prior keypoint proposals we follow \citet{jakab2018unsupervised} and use their proposed thin-plate-spline (TPS) augmentation for the prior image $x_p$, which is split to patches of size $8\times 8$ ($D=8$), resulting in a total of $K_p=256$ which are then filtered to $ L=50$ KP proposals. We follow \citet{thewlis2017unsupervised1, thewlis2017unsupervised2, jakab2018unsupervised} and use the unsupervised keypoints to regress from $K= \{25, 30, 50 \}$ to the annotated keypoints in the MAFL dataset. The linear regressor is learned without the bias term. For the regressor input, we experiment with using just the mean $\mu$ as features (deterministic KP, as in all previous works) and using the mean $\mu$ and the log-variance $\log(\sigma^2)$ as features. In Table \ref{tab:face-sota} we report the results in terms of the standard MSE normalized by the inter-ocular distance expressed as a percentage. It can be seen that \textbf{our method improves upon the SOTA in unsupervised keypoint discovery}.
Complete results, hyperparameters and an extended comparison can be found in Appendix \ref{apndx:res}.

\textbf{Information from uncertainty:} to test whether the learned variance of each particle contains meaningful information, we performed two experiments. First, we trained our model with $K=25$ particles and used the mean $\mu$ and the log-variance $\log(\sigma^2)$ as features for the supervised regression task described above. As seen in Table \ref{tab:face-sota}, we outperform \citet{jakab2018unsupervised} with $K=50$, even though the number of input features to the regressor is the same. In the second experiment, we used the model trained on $K=30$ keypoints and chose the $10$ keypoints with the highest variance and $10$ keypoints with lowest variance, and used their means $\mu$ as the input features to the regressor. For the low-variance batch, we report an error of 5.75\%, while for the high-variance batch the error was 7.54\%. The results from both experiments indicate that \textbf{the posterior variance is related to uncertainty in the location of the keypoints, and can be useful for decision making in downstream tasks}. We further illustrate the connection between the location of the particles and their variance in Appendix \ref{apndx:uncertainty}.

\begin{table}
    \setlength{\tabcolsep}{4pt}
    \centering
    \begin{small}
    \begin{tabular}{@{}llc@{}}
    \toprule
    \multicolumn{1}{c}{Method}                    & $K$  & MAFL        \\ \midrule
    Zhang~\citep{zhang2018kp}            & 30   & 3.16                           \\
    KeyNet~\citep{jakab2018unsupervised}   & 30 &   2.58                            \\
                                            & 50 & 2.54                             \\ \midrule
    Ours                                    & 25   & 2.87                         \\
                                            & 30   & \textbf{2.56}                            \\
                                            & 50   & \textbf{2.43}                          \\ \midrule
    Ours+ (with log-variance)               & 25   & \textbf{2.52}                         \\
                                            & 30   & 2.49                            \\
                                            & 50   & 2.42                          \\ \bottomrule
    \end{tabular}
    \end{small}
    \caption{{\bf Comparison with state-of-the-art on MAFL.} $K$ is the number of unsupervised landmarks. We report the MSE in \% between predicted and ground-truth (lower is better) obtained from the \texttt{Masked}-model.}
    \label{tab:face-sota}
    \vspace{-2em}
\end{table}

\subsection{Scene Decomposition and Image Manipulation}
\label{subsec:manip}
A hallmark of latent space generative models is the ability to change the image by controlling the latents~\citep{daniel2020soft, karras2020analyzing}.
Our model allows to modify the image in an intuitive way, by simply moving around the particles.

\begin{figure*}
     \centering
     \includegraphics[width=\textwidth]{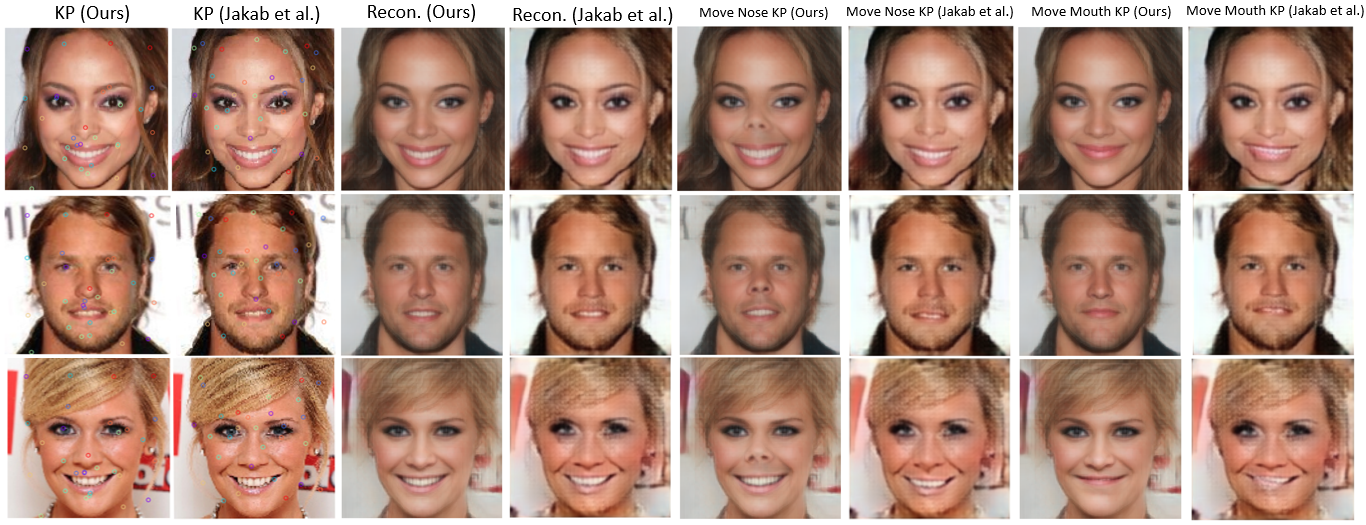}
     \vspace{-2em}
        \caption{Image manipulation comparison with KeyNet~\citep{jakab2018unsupervised}. We visualize the keypoints learned by each model, the reconstruction, and the effect that moving keypoints on the nose and the mouth has on the output image.}
        \label{fig:vs_jakab_1}
\end{figure*}

We first demonstrate the latent control with the \texttt{Masked} model,  trained with $K=30$ particles on CelebA~\citep{liu2015faceattributes} (cf.~Section \ref{subsec:supervised_kp}), and compare with KeyNet ~\citep{jakab2018unsupervised}, using their published pre-trained model with $K=30$ keypoints~\citep{keynet18code}. 
To perform manipulation, we visually locate the keypoints on the nose and mouth, slightly change their coordinates (leaving their features the same), and decode a new image. For a fair comparison, as keypoints differ between the models, we manually chose and moved the keypoints of \citet{jakab2018unsupervised} to produce the most visually pleasing results. 
In Figure \ref{fig:vs_jakab_1} we show the keypoints detected by each model, the reconstruction, and the resulting reconstruction after performing the above manipulation. As our latent space is structured to disentangle position and appearance, changing the position only affects the respective area of the particle by performing a smooth interpolation. KeyNet, on the other hand, has limited controllability as the latent space is only represented by the keypoints, and the features are propagated from the encoder, resulting in a blurry area near the keypoint position. Note that the manipulation in our model had a semantic effect -- moving the lip particle closed the mouth and hid the teeth, while moving the nose particle up exposed the nostrils.

As can be observed in Figure \ref{fig:vs_jakab_1}, several particles may be located in a small region (e.g., multiple particles near the nose). This is an attribute of the model, when the number of keypoints chosen is larger than the natural number of keypoints required to represent the variation in the data.  In such a case, not all particles have `control': manipulating them will have no effect on the reconstructed image. Interestingly, we found that controllable, or salient, particles are assigned lower uncertainty, and a simple filtering heuristic can be used to automatically select the top-$K$ salient particles, as we show in the supplementary material.

Next, we train our \texttt{Object} model on two multiple-object datasets: CLEVRER~\citep{yi2019clevrer} dataset and Traffic -- a self-collected traffic camera dataset. The CLEVRER dataset is composed of 5-second (128 frames) video of rigid objects colliding, where each frame can contain up to 8 objects of various shapes and colors. For this dataset, we learn $K=10$ particles with feature dimension $d=5$. Traffic is composed of 44,000 frames containing cars of different sizes and shapes. For this dataset, we learn $K=15$ particles with feature dimension $d=20$.
We emphasize that while these datasets contain videos, our method works on single images, and therefore ignores any temporal relation between the frames. We downscale the frames in both datasets to $128 \times 128$, use a glimpse size $S=32$, and do not use augmentations, i.e., $x_p=x$.

\begin{figure*}
     \centering
     \includegraphics[width=0.7\textheight]{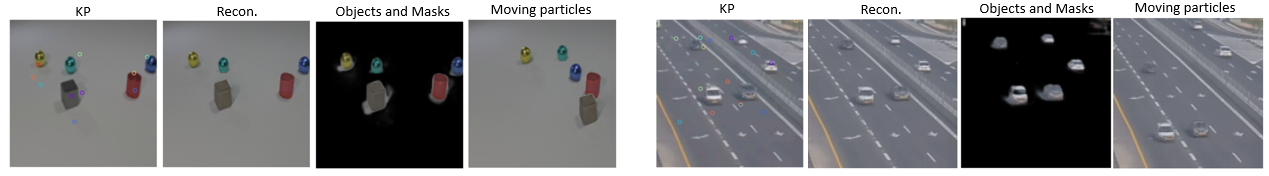}
        \caption{Scene decomposition and manipulation with  \texttt{Object} model. Left - CLEVRER, right - Traffic. We show the detected particles, the reconstructed images, the objects and masks -- output of the glimpse decoder, and image manipulation based on moving the particles.}
        \label{fig:obj_manip}
\end{figure*}

In Figure \ref{fig:obj_manip} we visualize a sample of the detected KP, the reconstructed images, the detected objects and their masks, and image manipulation by moving the KP (features remain the same).
Evidently, \textbf{our method can learn to decompose scenes with a varying number of objects of different shapes and sizes, and allows for particle-based manipulation of scenes where particles control objects}. 
Additional results can be found in Appendix \ref{apndx:scene_manip_res}\footnote{We have implemented a graphical use interface (GUI) for manipulating the images, please visit \url{https://taldatech.github.io/deep-latent-particles-web/} for videos. }.

We finally remark that empirically, we found that applying the \texttt{Masked} model in object-based scenes resulted in worse reconstructions, where objects were reconstructed as blurry blobs. 
Alternatively, using the \texttt{Object-based}  model
for non-object-based scenes reduced the manipulation ability. We noticed
that the model tended to assign `objects' to high-contrast parts of the image, such as the hairline and eyebrows, and ignored smoother parts such as the nose. We believe this is since high-contrast objects are easier to reconstruct using the alpha channel approach in the \texttt{Object-based}  model.

\subsection{Particle-based Video Prediction}
\label{sec:video}
The image manipulation results above suggest that our particles effectively control scene generation. We capitalize on this observation, and suggest to use particles for video prediction. Recall that particles are learned per image; our idea is to also learn a predictor for the \textit{temporal change} in particles, from video sequence data. A natural predictor that can exploit the disentangled position and appearance is a Graph Convolutional Network~(GCN,~\citealt{kipf2016gcn}), where each particle is a node, and connectivity is based on the Euclidean distance between particle positions. 
In this work, for simplicity, we chose a deterministic prediction model. Stochastic prediction, such as in \citet{minderer2019unsupervised}, can also be used with our model, and we leave that to future work.

We demonstrate our approach on the Traffic dataset, using our DLP model from Section \ref{subsec:manip}. We employ a 2-layer Gated GCN~\citep{bresson2017gatedgcn}
to predict the change in position $\Delta z_p$ and appearance features $\Delta z_a$ for each particle. To reduce drift in appearance, we constrain the maximal $\Delta z_a$ to a small value. The GCN is trained to predict particles of two consecutive frames $[t,t+1]$ from particles in two previous frames $[t-1,t]$. 
Since DLP is trained per-image, particles in consecutive frames do not necessarily match. Therefore, we do not have a ground truth for $\Delta z_p, \Delta z_a$.\footnote{In principle, the Chamfer distance can be used to resolve this, but in practice it only worked well for short horizon predictions.} Instead, we decode an image from the predicted particles, and train the GCN to minimize the perceptual loss~\citep{hoshen2019non} with the ground truth future frame. Video predictions are generated by rolling out the GCN, starting from the particles of the first two images in the video, and using the decoder to generate images from predicted particles. We provide the complete technical description of our method in the supplementary.
As a baseline, we trained the method of \citet{minderer2019unsupervised}, where keypoints are learned with KeyNet~\citep{jakab2018unsupervised}, on the Traffic dataset with the recommended hyperparameters~\citep{minderer19code}.

\begin{figure*}[h]
     \centering
     \includegraphics[width=\textwidth]{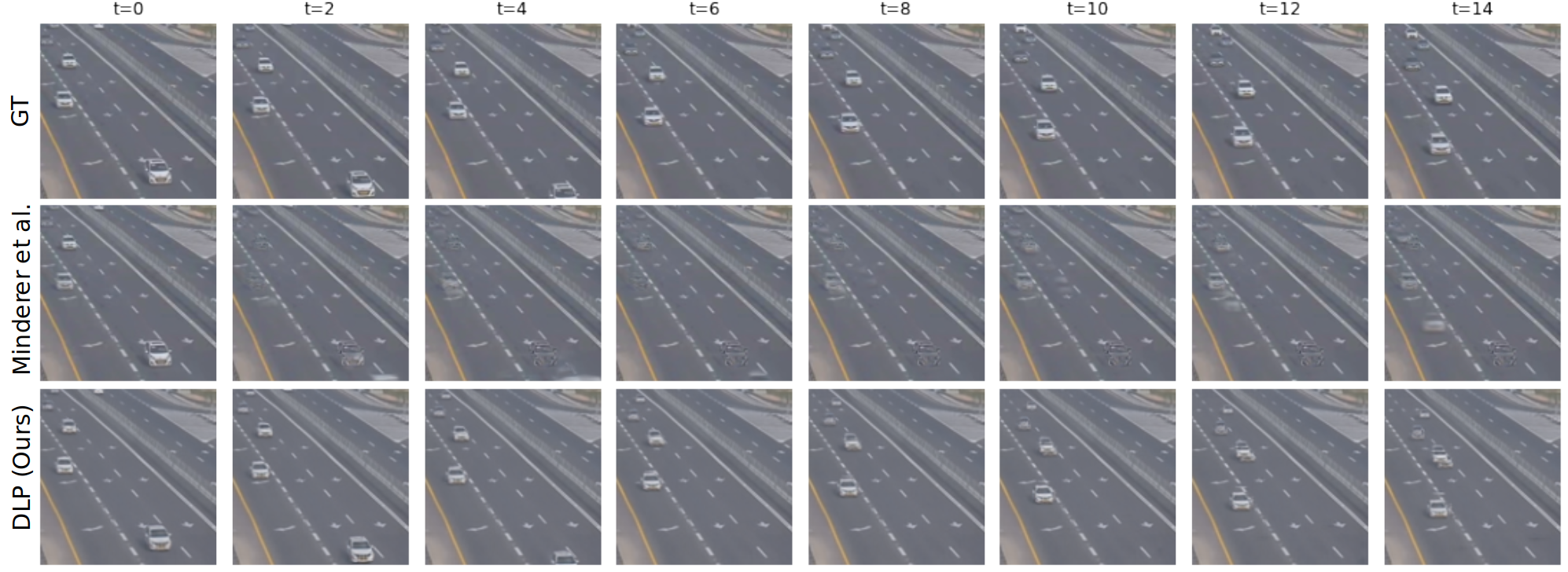}
        \caption{Video prediction on the Traffic dataset. We use the pre-trained \texttt{Object} model to provide the particle-representation and a GNN to predict the temporal change in particles. Top - ground truth, middle - \citet{minderer2019unsupervised} prediction, bottom - DLP (ours) prediction.}
        \label{fig:video_pred}
\end{figure*}

As can be seen in Figure \ref{fig:video_pred}, our approach produces sharp predictions, even for significantly longer horizons than trained on. 
We see this as a promising approach to video prediction, which is often prone to blurriness~\cite{lee2018savp, yan2021videogpt, hafner2020dreamer2}. We provide more details and results in the appendix. 
Note that the video prediction quality of the \citet{minderer2019unsupervised} baseline is significantly worse. In the appendix, we verify that the baseline obtains good reconstructions per single frames.
We hypothesize that the reason for the poor video prediction is the \textit{varying number of objects} (cars) in the scene: in all of the experiments in \citet{minderer2019unsupervised}, the number of objects in the scene was fixed, and thus their recurrent neural network approach was reasonable. Our GNN approach can better account for a variable number of objects. Additionally, as the baseline uses KeyNet, it is less capable of manipulability, as we demonstrated above.

For more complex scenes 
such as CLEVRER, we found that our simple approach does not work well enough, as we describe in Section~\ref{sec:limits}.

\subsection{Ablative Analysis}
\label{sec:ablation}
We evaluate the importance of our novel components, the Chamfer-KL and SSM prior. 
The ablation of Chamfer-KL uses the conventional KL calculation performed by flattening the particle representation to a vector. For this comparison, we chose a model with the same number of prior and posterior keypoints, $K_p=K=30$.
For an ablation of the SSM-based conditional prior, we experimented with two alternative priors: (1) a Gaussian prior $\mathcal{N}(0, 0.1^2)$, similar to the standard VAE setting; and (2) prior keypoint proposals sampled uniformly $\mathcal{U}[-1, 1]$ instead of using the SSM.
We experiment with the supervised KP regression task using the \texttt{Masked}-model.
We run the training for 50 epochs and keep the rest of the hyperparameters similar to Section \ref{subsec:supervised_kp}. As shown in Table \ref{tab:face-ablv}, using \textbf{the Chamfer-KL and the SSM-based prior significantly improves the performance}.

\begin{table}
    \setlength{\tabcolsep}{4pt}
    \centering
    \begin{small}
    \begin{tabular}{@{}llc@{}}
    \toprule
    \multicolumn{1}{c}{Prior/ KL}                    & Standard KL  & Chamfer-KL        \\ \midrule
    Constant $\mathcal{N}(0, 0.1^2)$                 & 3.18         & 3.07                         \\
    Random $\mathcal{U}[-1, 1]$                      & 3.39         & 2.99                         \\
    SSM                                              & 3.24         & \textbf{2.65}                          \\ \bottomrule
    \end{tabular}
    \end{small}
    \caption{{\bf Ablation study.} The effect of the choice of prior and KL implementation on the supervised KP regression on the MAFL dataset. Reported results are the MSE in \% between predicted and ground-truth KP (lower is better). Setting is different than in Table \ref{tab:face-sota}, see text for full information.}
    \label{tab:face-ablv}
    \vspace{-1em}
\end{table}

\section{Limitations and Future Work}
\label{sec:limits}
We illustrate several limitations of our method, which we observed when trying to predict video on CLEVRER.

\textbf{A particle shared between two objects:} when the data contains objects that frequently appear together, a single particle can be assigned to more than one object. This is due to our single-image formulation -- there is no signal for our method to separate the two objects. We believe that our method can be extended to a sequential setting, using the prior $x_p$, which would potentially resolve this issue for objects that change position during the video. \textbf{Large objects and occlusions:} several particles are assigned to large objects. This can become a problem when objects get partially occluded, requiring drastic changes to the particle features. More expressive GCNs trained on longer horizons may potentially mitigate this problem. An alternative is to modify the model to better account for occlusions. \textbf{Changing background:} the proposed model will have difficulty when the background changes dramatically, such as when the camera is moving. Incorporating some background detection into our method is one direction for addressing this.

\section{Conclusion}
\label{sec:conc}

In this work we showed that the classical concept of keypoint detection can be viewed in the lens of deep generative modelling, by viewing the keypoints themselves as the latent variables in a variational autoencoder model. Beyond the elegance of the formulation, we showed that our method can generate SOTA results in keypoint discovery, and be used for intriguing image manipulations. 

Many questions remain. For example, extending the method to handle more complex video prediction, where objects change appearance dramatically, or occlude other objects. Another exciting direction is to leverage developments in VAEs, such as the vector-quantized VAE~\citep{razavi2019vqvqae} for improved performance. We also intend to explore the use of the uncertainty estimate in our model for decision making. More broadly, we are hopeful that this new connection between generative models and keypoint detection will spur up interesting developments in image representations.

\section{Acknowledgements}
\label{sec:ack}
The authors would like to thank Ron Mokadi for collecting and sharing the Traffic dataset used in this work. This work is partly funded by the Israel Science Foundation (ISF-759/19), the Open Philanthropy Project Fund, an advised fund of Silicon Valley Community
Foundation, and by the European Union. Views and opinions expressed are however those of the author(s) only and do not necessarily reflect those of the European Union or the European Research Council. Neither the European Union nor the granting authority can be held responsible for them.

\newpage
\bibliography{example_paper}
\bibliographystyle{icml2022}

\newpage
\appendix
\onecolumn
\section{Extended Architecture Details}
\label{apndx:arch}
\paragraph{Encoders:} all of the encoders described in this work follow a similar scheme to \citet{jakab2018unsupervised}. Unless mentioned otherwise, input images are assumed to be of $128\times128$ resolution. The position encoder is composed of a CNN with convolutional blocks where each block contains a convolutional layer, followed by Batch Normalization and \texttt{ReLU} activation. Downsmapling is performed by using strided ($s=2$) convolutional layers with `replication` padding. The channels of each convolutional block are $[32, 64, 128, 256]$ and the feature maps are of shape $16\times16\times 256$. These maps are flattened and fed into 3-layer fully-connected network with hidden layers of size $[256, 128]$, each activated with \texttt{ReLU}, outputting $K\times4$ position values, reshaped to $[\mu, \log(\sigma^2)]$. The prior encoder, operating on patches of sizes $16\times16$ or $32\times32$, is composed of a CNN with a similar structure to the position encoder with channels $[16, 32, 64]$ followed by a spatial softmax (SSM) block. Finally, the glimpse encoder has similar structure as the prior encoder, but the SSM is replaced with a FC network similar to the one in the position encoder.

\paragraph{$\dgraph$ - PointNet++:} to decode feature maps from the latent particles, we use a PointNet++~\citep{qi2017pointnet++} implemented as a GNN~\cite{gnn09}. First, a KNN graph with $K=10$ is built from the position of the particles. This graph is then processed by a 4-layer PointNet++ layers, composed of 1-D convolution, \texttt{ReLU} activation and Batch Normalization, with channels $[64, 128, 256, 512]$. The output of the convolutional blocks is then max-pooled to produce a vector which is reshaped to $[K, 8, 8]$ maps, where $K$ is the number of particles. This is implemented efficiently with the \texttt{torch geometric} library~\citep{feylensen19tgeom}. Finally, these maps are upsampled with a convolutional layer to $[16, 16, K]$ maps.

\paragraph{$\dupsample$ - Upsampling CNN:} this component takes in a concatenation of feature maps; For the \texttt{Masked} model, the concatenation is of the Gaussian heatmaps $\Phi_{heatmap}$, masked feature maps from the encoder $\Phi_{enc}$ and masked feature maps from $\Phi_{graph}$. The final shape of the input to $D_{upsample}$ is $[16, 16, K\times3]$. For the \texttt{Object} model, the concatenation is of the Gaussian heatmaps $\Phi_{heatmap}$ and feature maps $\dgraph$. The final shape of the input to $D_{upsample}$ is $[16, 16, K\times2]$.
The upsampling CNN network has a symmetric structure to the CNN in the position encoder, but in reverse, namely, the CNN channels are: $[256, 128, 64, 32]$. Finally, the output is aggregated with a $1\times 1$ convolutional layer to produce a RGB image of shape: $[128, 128, 3]$.

\paragraph{$\dglimpse$ - Glimpse decoder:} this component takes in a latent vector which is mapped with a 2-layer fully-connected network with 256 hidden units each and \texttt{ReLU} activated to a vector of size $8\times8\times32$. This vector is reshaped to feature maps of shape $[8, 8, 32]$ that are upsampled with a 2-layer CNN (similar blocks as before) with $64$ channels in each layer to maps of shape $[S, S, 64]$, where $S$ is the glimpse size. This output is aggregated with a $1\times1$ convolutional layer and activated with a \texttt{Sigmoid} activation to produce RGBA patches of shape $[S, S, 4]$.

\section{Extended Implementation and Training Details}
\label{apndx:impl}
In this section, we describe important implementation and training details for convergence of our method.
\paragraph{Training objective:} the complete training objective follows the $\beta$-VAE~\citep{higgins2017beta} formulation: $$ \mathcal{L} =  \mathcal{L}_{rec}(x, \Tilde{x}) +\beta_{CKL}CH-KL(q_{\phi}(z_p|x)\Vert p_{\psi}(z|x)) + \beta_{KL} KL(q_{\phi}(z_a|x) \Vert p(z)).$$ We found it crucial to balance the two KL terms, where usually $\beta_{CKL} > \beta_{KL}$. We report the exact values for each task in Appendix \ref{apndx:hyp}.

\paragraph{Initialization, LR scheduling and latent activation:} in our implementation, we initialized the convolutional layers with values from $\mathcal{N}(0, 0.01)$. We used a general multi-step learning scheduler that decreases the learning by $0.5$ in each milestone. The milestones are reported in Appendix \ref{apndx:hyp}. Moreover, to constrain the values of the particles' position to be in the range of $[-1, 1]$, we used a \texttt{TanH} activation on $\mu$.

\paragraph{Binary masks from Gaussian heatmaps:} for the \texttt{Masked} model, we use a binary mask $M_u(x)$ created from the Gaussian as follows: $$M_u^i(x) = \begin{cases} 1, & \Phi_{heatmap}^i(x) \geq \tau \\ 0, & \text{else} \end{cases},$$ for some threshold $\tau$ ($\tau = 0.2$ in our experiments). Now, the input of $\dupsample$ is a concatenation of $\Phi_{heatmap}$, $M \odot \dgraph$ and $(1 - M) \odot \Phi_{enc}(x)$.

\paragraph{Warm-up, noisy masks and transparency variable:} for the \texttt{Object} model, we found it beneficial to have a short warm-up stage of 2 epochs where only the glimpse encoder and glimpse decoder are training to encode and decode patches. This warm-up step prevents $\dupsample$ to take full responsibility of the reconstruction and ignore $\dglimpse$. Moreover, at the beginning of the training (5-10 epochs) we add a small Gaussian noise $\mathcal{N}(0, 0.01)$ to the alpha channel of the decoded patches following \citet{smirnov2021marionette}. This additional step encourages learning sharper object masks. Finally, following \citet{smirnov2021marionette}, we learn an additional `transparency` parameter for each particle $z_{on}\in [0, 1]$ which is multiplied by the decoded glimpse prior to stitching the final reconstructed image. This implemented by extending the output dimension of the FC layer in $\Phi_{enc}$ and using \texttt{Sigmoid} activation for that variable.

\paragraph{Frozen prior network} Interestingly, even though the prior network can be trained (the SSM is differentiable), we found that in some datasets (e.g. Traffic and CLEVRER), keeping the network frozen, with the initial random weights, worked better. This is inline with the findings of \citet{frankle2020training}.

\paragraph{Stitching the image:} we follow a layer-wise approach~\citep{smirnov2021marionette} and describe our stitching algorithm in Algorithm \ref{alg:stitch}.

\begin{algorithm}[tb]
  \caption{Stitching Algorithm}
  \label{alg:stitch}
\begin{algorithmic}
  \STATE {\bfseries Input:} alpha maps ${a}_{i=1}^K$,  RGB maps ${r}_{i=1}^K$, background map $b$
  \STATE Initialize $currMask = a[1]$, $masks = []$.
  \FOR{$i=2$ {\bfseries to} $K$}
  \STATE $availableSpace = 1.0 - currMask$
  \STATE $currMaskTmp = Min(availableSpace, a[i])$
  \STATE Append $currMaskTmp$ to $masks$
  \STATE $currMask = currMask + currMaskTmp$
  \ENDFOR
  \STATE $alphaMask \leftarrow$ Sum($masks$)
  \STATE $rec = (1 - alphaMask) * bg + masks * r$
  \STATE Return $rec$
\end{algorithmic}
\end{algorithm}

\section{Datasets}
\label{apndx:data}
In this section we provide a detailed description of the datasets we used throughout this paper.

\paragraph{CelebA~\cite{liu2015faceattributes}:} this dataset contains 200k images of celebrity faces which are cropped and resized to $128\times128$ following \citet{jakab2018unsupervised, thewlis2017unsupervised1}. The dataset provides
annotations for 5 facial landmarks — eyes, nose and mouth corners, which are not required during training. As per the setting of \cite{jakab2018unsupervised, thewlis2017unsupervised1}, the MAFL~\citep{zhang2014facial} test-set is excluded from training.

\paragraph{CLEVRER~\cite{yi2019clevrer}:} this dataset contains 20,000 synthetic videos of moving and colliding objects, separated to 10,000 train video, 5,000 validation videos and 5,000 test videos, where each video is 5 seconds long and contains 128 frames with resolution $480 \times 320$. In our work, we resize the frames to $128 \times 128$ and use a subset of the frames created by skipping every second frame.

\paragraph{Traffic:} a self-collected traffic camera dataset composed of 44,000 frames resized to $128\times128$ containing cars of different sizes and shapes, where we take the first 90\% of the frames for train and the rest for evaluation.

\section{Hyperparameters Details}
\label{apndx:hyp}
In this section we provide the complete set of hyperparameters used for the experiments in this work. The shared hyperparameters between all of the experiments are described below, and the rest can be found in Table \ref{tab_apndx:hyp}.

\paragraph{Shared hyperparameters:} all models were trained with an initial learning rate of $2e-4$ and a a batch size of $32$ per GPU (we used 1 to 4 GPUs). For all datasets, we used a multi-step learning rate scheduler with the following milestones (in epochs): $[30, 60]$ with learning rate decreasing by $0.5$ on each milestone. The warm-up stage described in Appendix \ref{apndx:impl} was only used for the \texttt{Object} model, where we used 1 warm-up epoch for CLEVRER and 2 for Traffic, and the number of `noisy masks` epochs was 5 times the warm-up epochs--5 and 10 for CLEVRER and Traffic, respectively.

\begin{table}
    \setlength{\tabcolsep}{4pt}
    \centering
    \begin{tabular}{cccccccccccccc}
    \toprule
    dataset                      & Model            & $K$ & $K_p$ & $L$ & $\beta_{CKL}$ & $\beta_{KL}$ & prior patch size & glimpse size $S$ & feature dim $d$ & epochs       \\ \midrule
    CelebA                      & \texttt{Masked}   & 10  &    256    &  15  &    20         &  20 $*$ 0.001 &       8          &    16     &      30           & 90       \\
    CelebA                      & \texttt{Masked}   & 25  &    256     & 30    &     30      &  30 $*$ 0.001 &        8          &    16  &        10           & 90    \\
    CelebA                      & \texttt{Masked}   & 30  &   256      &  50  &      40      &  40 $*$ 0.001 &        8          &  16       &      10          & 90         \\
    CelebA                      & \texttt{Masked}   & 50  &    256     &  50  &       50      &  50 $*$ 0.001 &        8          &   16   &        10          & 90         \\ \midrule
    Traffic                     & \texttt{Object}   & 15  &     64    &  64    &      30          & 30  $*$ 0.001 &        16     &        32          &  20         &  100          \\
    CLEVRER                     & \texttt{Object}   & 10  &    64     &  64    &     40           & 40 $*$ 0.001 &       16        &       32          &    5         & 120          \\ \bottomrule
    \end{tabular}
    \caption{Detailed hyperparameters used for the various experiments in the paper.}
    \label{tab_apndx:hyp}
\end{table}

\section{Complete Results}
\label{apndx:res}
In this section, we provide extended results for the various tasks we presented in the main paper.

\subsection{Supervised Regression on Face Landmarks}
\label{apndx:supervised_kp}
In Table \ref{tab_apndx:face-sota} we present the complete set of results for the supervised KP regression task. It can be seen that our method outperforms the supervised and unsupervised benchmark for $K=\{30, 50\}$, and combined with the uncertainty information and the learned features, the results are further improved. It is worth noting that simply learning features without the notion of location results in bad performance as reported by \citet{jakab2018unsupervised}, stressing that the learned features are only informative when learned with respect to their position information.
\begin{table}
    \setlength{\tabcolsep}{4pt}
    \centering
    \begin{tabular}{@{}llc@{}}
    \toprule
    \multicolumn{1}{c}{Method}                    & $K$  & MAFL        \\ \midrule
    \multicolumn{3}{c}{Supervised}                                           \\
    RCPR~\citep{burgos2013robust}                 &      & {--}                         \\
    CFAN~\citep{zhang2014coarse}                  &      & 15.84                         \\
    Cascaded CNN~\citep{sun2013deep}              &      & 9.73                          \\
    TCDCN~\citep{zhang2015learning}                       &      & 7.95                           \\
    MTCNN~\citep{zhang2014facial}                 &      & 5.39                         \\ \midrule
    \multicolumn{3}{c}{Unsupervised / Self-supervised}                                           \\
    Thewlis~\citep{thewlis2017unsupervised1}          & 30   & 7.15                         \\
                                                  & 50   & 6.67                         \\
    Thewlis~\citep{thewlis2017unsupervised2}(frames) & {--} & 5.83                          \\
    Shu~\citep{shu2018deforming}         & {--} & 5.45                            \\
    Zhang~\citep{zhang2018kp}            & 10   & 3.46                           \\
                                           & 30   & 3.16                           \\
    Wiles~\citep{wiles2018self}              & {--} & 3.44                           \\ 
    KeyNet~\citep{jakab2018unsupervised}   & 10 &   3.19                            \\
                                            & 30 & 2.58                             \\
                                            & 50 & 2.54                             \\ 
    Lorenz~\citep{lorenz2019unsupervised}   & 10 &   3.24                            \\
    Dundar~\citep{dundar2021unsupervised}   & 10 &   2.76                      \\ \midrule
    Ours                                    & 10   & 3.87                         \\
                                            & 25   & 2.87                            \\
                                            & 30   & 2.56                            \\
                                            & 50   & 2.43                          \\ \midrule
    Ours+ (with log-variance)                   & 10   & 3.12                         \\
                                            & 25   & 2.52                            \\
                                            & 30   & 2.49                            \\
                                            & 50   & 2.42                          \\ \midrule
    Ours++ (with learned features)          & 10   & 2.98                         \\
                                            & 25   & 2.42                            \\
                                            & 30   & 2.36                            \\
                                            & 50   & 2.39                         \\ \bottomrule
    \end{tabular}
    \caption{{\bf Comparison with state-of-the-art on MAFL.} $K$ is the number of unsupervised landmarks. Reported results are the MSE in \% between predicted and ground-truth (lower is better).}
    \label{tab_apndx:face-sota}
\end{table}

\subsection{Uncertainty Information Analysis}
\label{apndx:uncertainty}
In this section, we demonstrate a visual connection between the location of each particle and its learned variance.
Each keypoint $u_k$ is defined by the Gaussian parameters $(\mu_k, \sigma_k)$, where $\sigma_k^2$ can be interpreted as the variance in the location of this keypoint. Intuitively, for common patterns in the data, 
we should expect the variance to be small.
Accordingly, we define the per-keypoint uncertainty as follows: $$V(u_k) \doteq \sum_i \log(\sigma_{k_i}^2),$$ where $\sigma_{k_i}$ is the standard deviation in the $i^{th}$ axis (i.e., the $x$ and $y$ coordinates) of $u_k$.\footnote{One may also consider the variance in the features for each particle. However, to illustrate our idea of disentangling position from appearance, we only consider position uncertainty.}

\begin{figure*}[h]
     \centering
     \includegraphics[width=\textwidth]{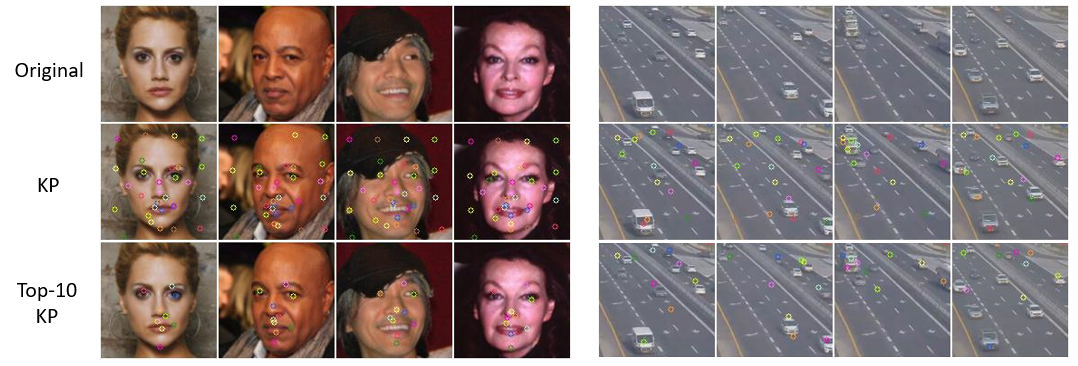}
        \caption{Information from uncertainty. We show the $K=30$ particles (second row) learned from two models and top-$10$ particles with the highest confidence (third row): (1) \texttt{Masked} model trained on CelebA (left); and (2) \texttt{Object} model trained on Traffic (right).}
        \label{fig:top_kp}
\end{figure*}

To test our hypothesis, we use two trained models: (1) \texttt{Masked} model from Section \ref{subsec:supervised_kp} on $128\times 128$ face images from CelebA~\citep{liu2015faceattributes} and (2) \texttt{Object} model from Section \ref{subsec:manip} on the Traffic dataset. The \texttt{Masked} was trained with $K=30$ particles and the \texttt{Object} model with $K=15$. In Figure \ref{fig:top_kp} we plot the $K$ keypoints learned by our model and display the top-$10$ keypoints with highest confidence. It can be seen that the keypoints with the highest confidence lie on locations that are common across the dataset (i.e., eyes, nose and mouth) while the rest lie in regions of higher variability (e.g., hair and background).

\subsection{Scene Decomposition and Image Manipulation}
\label{apndx:scene_manip_res}
\begin{figure*}
     \centering
     \includegraphics[width=\textwidth]{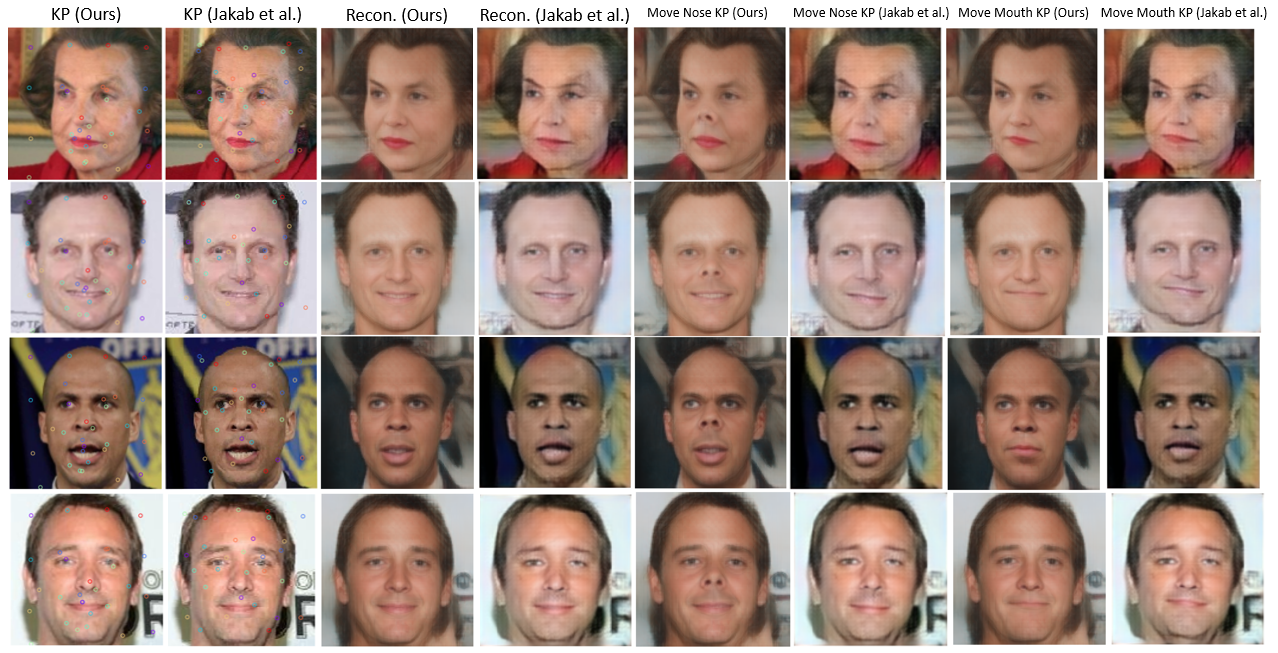}
        \caption{Image manipulation comparison with KeyNet~\citep{jakab2018unsupervised}. We visualize the keypoints learned by each model, the reconstruction, and the effect that moving keypoints on the nose and the mouth has on the output image.}
        \label{fig:vs_jakab_apndx}
\end{figure*}

We provide more results for the different experiments described in Section \ref{sec:exp}. First, we compare image manipulation of the \texttt{Masked} model with KeyNet~\citep{jakab2018unsupervised} on CelebA in Figure \ref{fig:vs_jakab_apndx}. The experimental setting, where both models learn $K=30$ keypoints, is the same as in Section \ref{subsec:manip}.

\begin{figure*}
     \centering
     \includegraphics[width=0.7\textwidth]{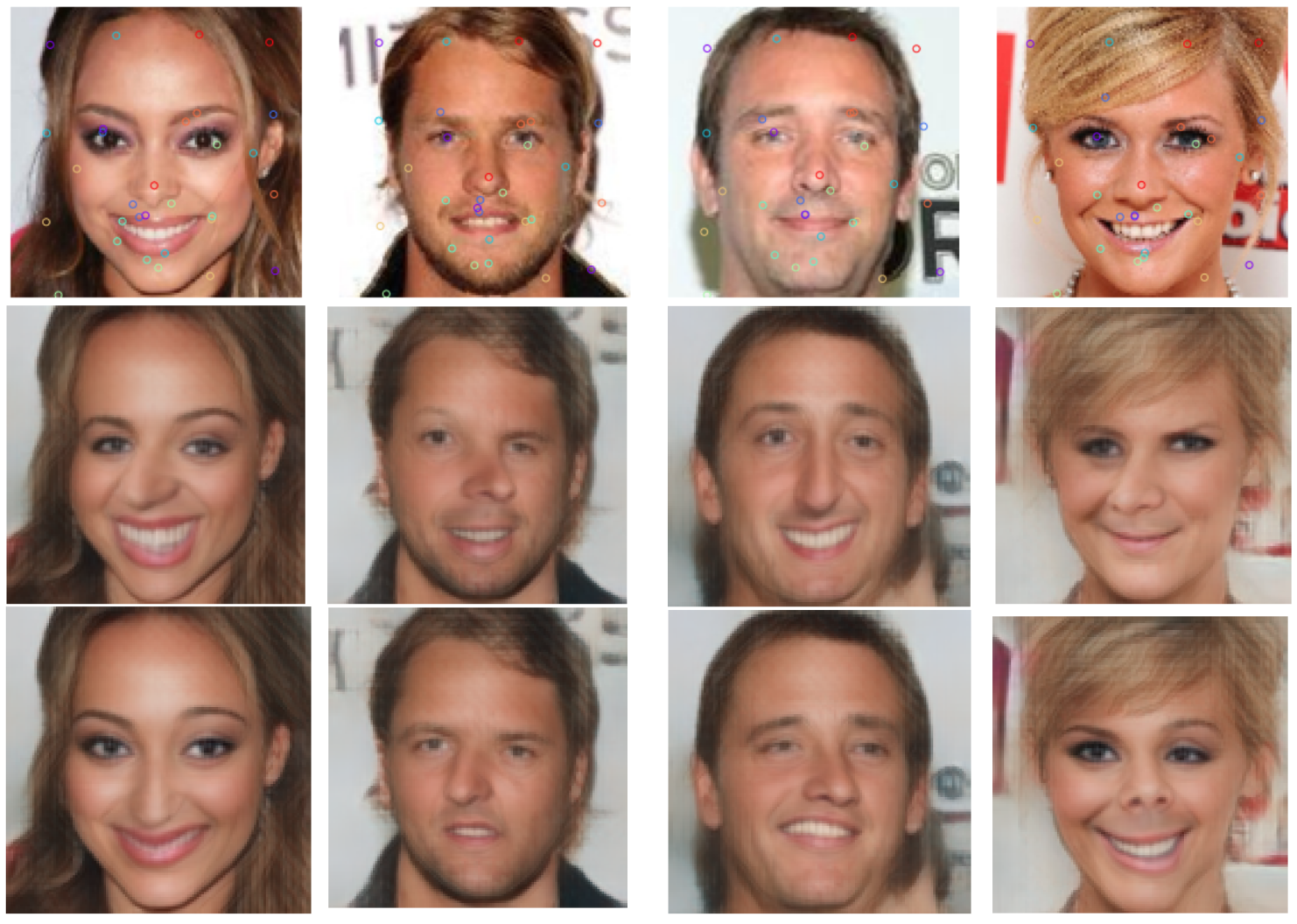}
        \caption{Image manipulation comparison with KeyNet~\citep{jakab2018unsupervised}. We visualize the effect of moving keypoints in the face area.}
        \label{fig:celab_manip_apndx}
\end{figure*}

In Figure \ref{fig:celab_manip_apndx} we provide more manipulations on CelebA produced by moving the particles in the face area.

\begin{figure*}
     \centering
     \includegraphics[width=0.5\textwidth]{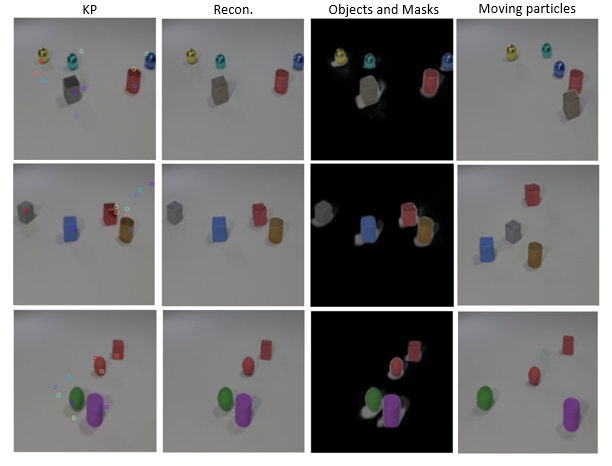}
        \caption{Scene decomposition and manipulation with \texttt{Object} model on CLEVRER. We show the detected particles, the reconstructed images, the objects and masks -- output of the glimpse decoder, and image manipulation based on moving the particles.}
        \label{fig:obj_manip_clevrer}
\end{figure*}

\begin{figure*}
     \centering
     \includegraphics[width=0.5\textwidth]{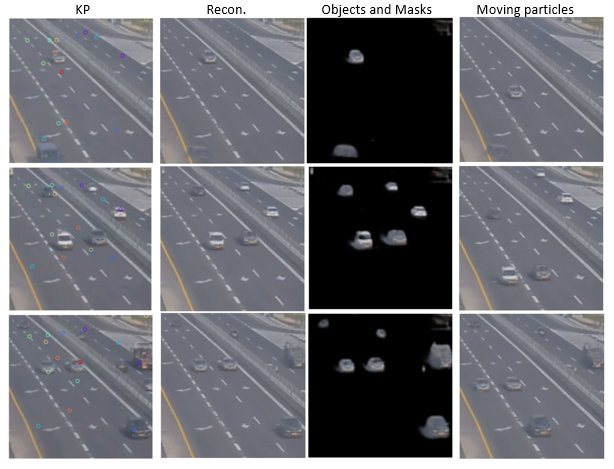}
        \caption{Scene decomposition and manipulation with \texttt{Object} model on Traffic. We show the detected particles, the reconstructed images, the objects and masks -- output of the glimpse decoder, and image manipulation based on moving the particles.}
        \label{fig:obj_manip_traffic}
\end{figure*}

Next, we provide extended results of our \texttt{Object} model. In Figures \ref{fig:obj_manip_clevrer} and \ref{fig:obj_manip_traffic} we visualize the detected particles, the reconstructed images, the objects and masks -- output of the glimpse decoder, and image manipulation based on moving the particles on CLEVRER and Traffic, respectively.

\subsection{Video Prediction}
\label{apndx:video_pred_res}
\begin{figure*}[h]
     \centering
     \includegraphics[width=\textwidth]{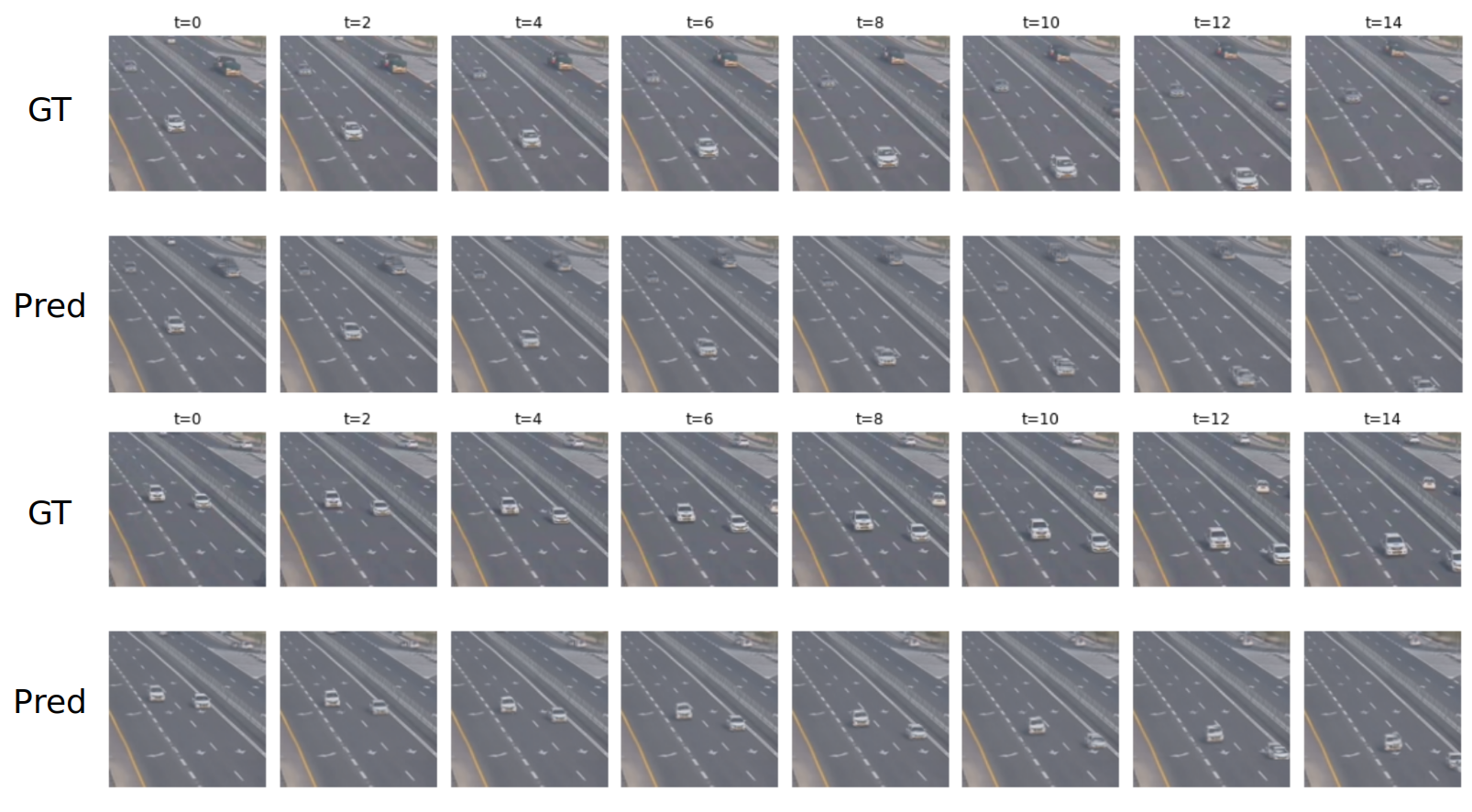}
        \caption{Video prediction on the Traffic dataset. We use the pre-trained \texttt{Object} model to provide the particle-representation and a GNN to predict the temporal change in particles. Top - ground truth, bottom - prediction.}
        \label{fig:video_pred_long_1}
\end{figure*}

\begin{figure*}[h]
     \centering
     \includegraphics[width=\textwidth]{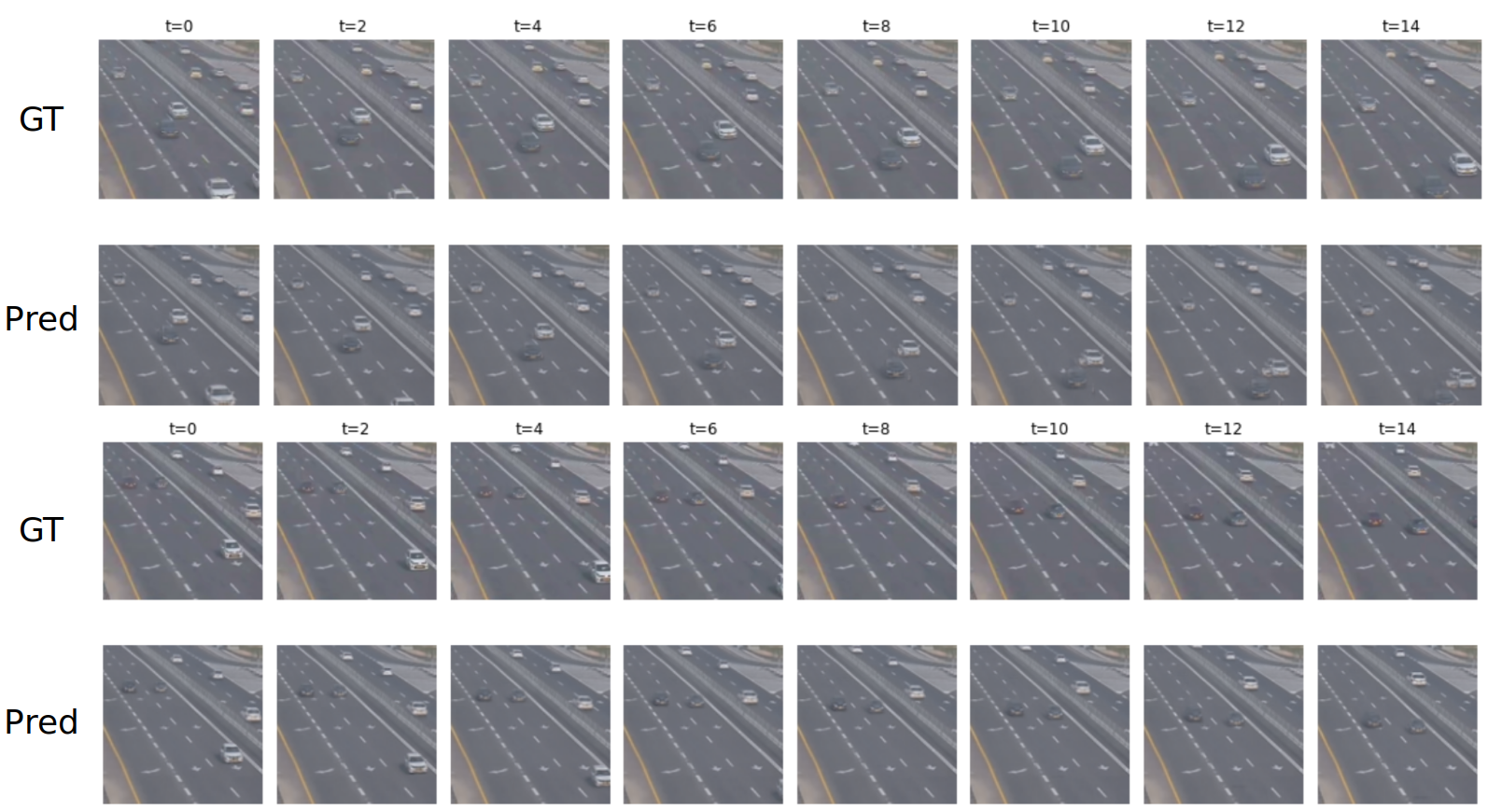}
        \caption{Video prediction on the Traffic dataset. We use the pre-trained \texttt{Object} model to provide the particle-representation and a GNN to predict the temporal change in particles. Top - ground truth, bottom - prediction.}
        \label{fig:video_pred_long_2}
\end{figure*}

We present more results for the video prediction experiment we presented in Section \ref{sec:video}. Video predictions are generated by rolling out the GCN, starting from the particles of the first two images in the video, and using the decoder to generate images from predicted particles.
As can be seen in Figures \ref{fig:video_pred_long_1} and \ref{fig:video_pred_long_2}, our approach produces sharp predictions, even for significantly longer horizons than trained on. For animated sequences please see supplementary material. In Figure \ref{fig:video_pred_minderer} we plot show single-frame reconstructions and video prediction of the method proposed in \citet{minderer2019unsupervised} trained on the Traffic dataset.

\begin{figure*}[h]
     \centering
     \includegraphics[width=\textwidth]{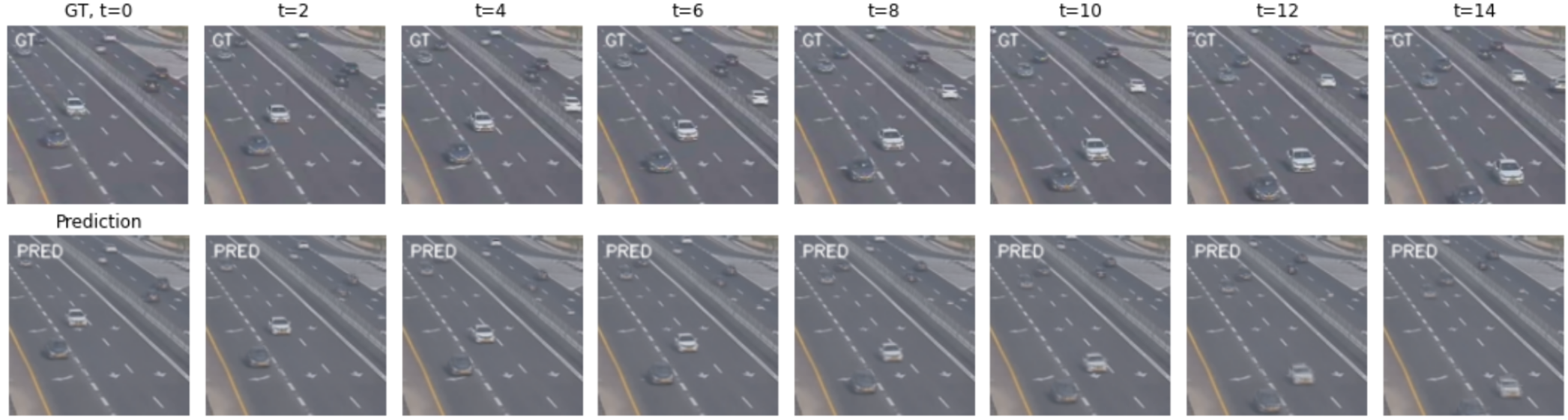}
        \caption{Video prediction on the Traffic dataset. We use the pre-trained \texttt{Object} model to provide the particle-representation and a GNN to predict the temporal change in particles. Top - ground truth, bottom - prediction.}
        \label{fig:video_pred_t}
\end{figure*}

\begin{figure*}[h]
     \centering
     \includegraphics[width=\textwidth]{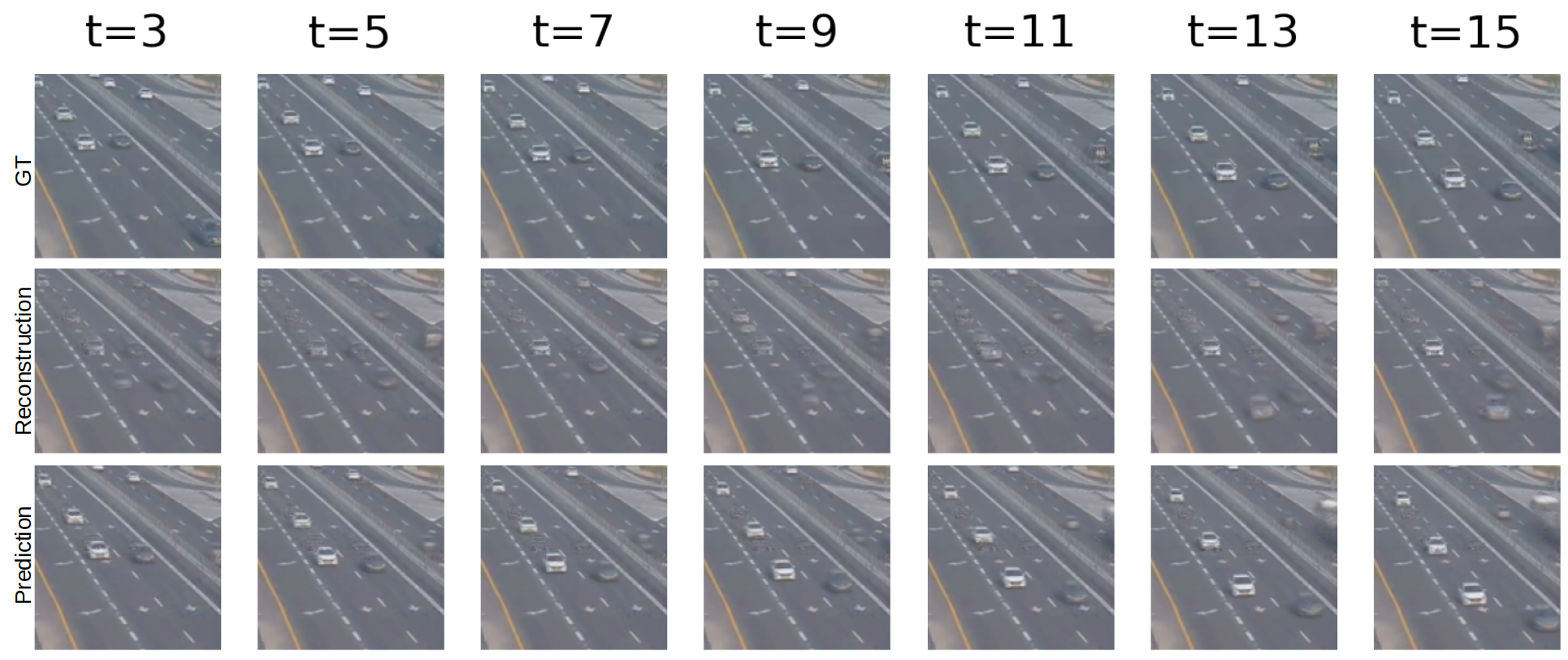}
        \caption{Video prediction baseline (Minderer et al.) on Traffic. Second row: reconstructions of KeyNet; third row: dynamics prediction by \citet{minderer2019unsupervised}}
        \label{fig:video_pred_minderer}
\end{figure*}

\section{Video Prediction Experiment Details}
\label{apndx:video}
In this section, we provide technical details of the video prediction experiment we described in Section \ref{sec:video}. The objective in this experiment is to learn a predictor for the temporal change of the particles, from video sequence data. Recent advances in graph neural networks (GNN,~\citealt{gnn09}) provide a natural framework to learn on graph-structured data. GNNs, such as Graph Convolutional Network~(GCN,~\citealt{kipf2016gcn}), can learn to extract details relevant to interaction between nodes in the graph. With DLP, we model each particle as a node and the connectivity between nodes is based on the Euclidean distance between particle positions. Furthermore, each node is described by the particle's features, namely, its position and appearance features.

To predict the change in position $\Delta z_p$ and appearance features $\Delta z_a$ for each particle, we employ a 2-layer Gated GCN~\citep{bresson2017gatedgcn}, with 128 hidden units each and activated with \texttt{ReLU}. To make the network more expressive, we add a 2-layer 1D convolutional network implemented as a shared MLP operating on each node separately, with 128 units each and activated with \texttt{ReLU}. The GCN is trained to predict particles of two consecutive frames $[t,t+1]$ from particles in two previous frames $[t-1,t]$ as follows: first, for each frame, we extract its respective particle-representation using a pre-trained DLP model and we concatenate a one-hot vector to its features indicating the time-step (e.g., $[1, 0]$ is concatenated to the features of the particle representation at time-step $t-1$). Then, we build a \textit{radius} ($r=0.2$) graph from the resulting set of nodes based on the Euclidean distance between the particles. This graph is fed into the GCN, which via the message-passing process outputs  $\Delta z_p$ and $\Delta z_a$ for the consecutive time-steps $[t,t+1]$. These updates are activated with a \texttt{TanH} activation and multiplied by constants $\gamma_p = 0.2$ and $\gamma_a=0.02$ to constrain the maximal $\Delta z_p$ and $\Delta z_a$, respectively. Finally, the $\Delta$ is added to the original particles (a residual connection) and the new particle is decoded back to the image space using the pre-trained DLP model. We note that the pre-trained DLP model stays frozen throughout the training process of the GCN. As DLP is trained per-image, particles in consecutive frames do not necessarily match, and the GCN is trained to minimize the perceptual loss~\citep{hoshen2019non} with the ground truth future frame. Video predictions are generated by rolling out the GCN, starting from the particles of the first two images in the video, and using the decoder to generate images from predicted particles. The GCN layes are implemented efficiently with the \texttt{torch geometric} library~\citep{feylensen19tgeom}.

\end{document}